%% file: IEEEICC.tex
\documentclass{article}
\usepackage{arxiv}

\usepackage[utf8]{inputenc} 
\usepackage[T1]{fontenc}    
\usepackage{booktabs}       
\usepackage{amsfonts}       
\usepackage{nicefrac}       
\usepackage{microtype}      
\usepackage{graphicx}
\usepackage{afterpage}
\usepackage{placeins}
\usepackage{cite}
\usepackage{amsmath,amssymb,amsfonts}
\usepackage{algorithmic}
\usepackage{graphicx}
\usepackage{textcomp}
\usepackage{mathtools}
\usepackage{subcaption}
\usepackage{multirow}
\usepackage{lipsum}
\usepackage{titlesec}
\usepackage{amsmath}
\usepackage[table,xcdraw]{xcolor}
\usepackage{array}
\usepackage{longtable}

\title{Deep Learning Approaches to Indoor Wireless Channel Estimation for Low-Power Communication}

\author{Samrah Arif*,
        M. Arif Khan*,
        and Sabih ur Rehman*
        \\ 
        {*School of Computing, Mathematics and Engineering, Charles Sturt University, Australia}}

\begin{document}
\maketitle
\begin{abstract}
	In the rapidly growing development of the Internet of Things (IoT) infrastructure, achieving reliable wireless communication is a challenge. IoT devices operate in diverse environments with common signal interference and fluctuating channel conditions. Accurate channel estimation helps adapt the transmission strategies to current conditions, ensuring reliable communication. Traditional methods, such as Least Squares (LS) and Minimum Mean Squared Error (MMSE) estimation techniques, often struggle to adapt to the diverse and complex environments typical of IoT networks. This research article delves into the potential of Deep Learning (DL) to enhance channel estimation, focusing on the Received Signal Strength Indicator (RSSI) metric - a critical yet challenging aspect due to its susceptibility to noise and environmental factors. This paper presents two Fully Connected Neural Networks (FCNNs)-based Low Power (LP-IoT) channel estimation models, leveraging RSSI for accurate channel estimation in LP-IoT communication. Our Model A exhibits a remarkable 99.02\% reduction in Mean Squared Error (MSE), and Model B demonstrates a notable 90.03\% MSE reduction compared to the benchmarks set by current studies. Additionally, the comparative studies of our model A with other DL-based techniques show significant efficiency in our estimation models.
\end{abstract}


\section{Introduction}
The advent of the IoT has become a pivotal advancement in digital connectivity and automation \cite{Ammar2019}. A critical element in the IoT ecosystem is reliable and efficient communication among devices, where channel estimation plays a vital role in maintaining signal transmission quality in dynamic environments. In this context, the RSSI emerges as a key metric for channel estimation due to its simplicity and minimal overhead \cite{Dinev2022}. RSSI-based channel estimation is vulnerable to environmental variables and noise, which affect link quality determination and device localisation. Therefore, it is imperative to have an accurate RSSI estimation for reliable LP-IoT wireless communication.
Conventional channel estimation techniques like LS and MMSE are often based on simplified linear models \cite{Samrah2024}. Therefore, such methods struggle to accurately capture the complex characteristics of real-world wireless channels, particularly in diverse and fluctuating environments \cite{Ye2018}\cite{Hu2021}.
Recent studies demonstrate the potential of Deep Learning (DL) in wireless channel estimation to overcome the limitations of traditional estimation methods \cite{Kim2023} \cite{Wu2020}. Inspired by biological neural networks, DL algorithms process and transmit information via interconnected artificial neurons, learn from input-output examples, and facilitate predictions on new data. The core of DL networks is the perceptron, which processes inputs through a neuron, computes a weighted sum with a bias, and applies a decision function to predict the output \cite{Saitoh2021}. 

While the emerging DL methodologies offer a transformative approach to wireless channel estimation, subsequent investigations continue to explore and refine these methods \cite{popescu2006}, focusing on the challenges in wireless channel estimation. In \cite{Raj2021}, the author investigates the application of Artificial Neural Networks (ANNs), also called FCNNs, for RSSI prediction in indoor environments, finding ANN's superiority to linear regression models in estimating the path loss exponent. However, the limited dataset size and substantial training and testing errors suggest further research to improve the ANN-based channel estimation model in indoor environments. Another research by the same author in \cite{Raj2023} introduces IndoorRSSINet, a DL model for predicting 2D RSSI maps using ray-tracing data, but faces limitations due to its reliance on synthetic data and the need for specialised data preprocessing. Further contributions in this domain include the DL-based approach in \cite{Thrane2020} and \cite{Katagiri2019}. In \cite{Thrane2020}, the authors present model-aided DL for signal strength prediction, while the author in \cite{Katagiri2019} presents a novel method for radio propagation prediction using a model classifier. While these investigations demonstrate significant contributions, they also indicate a need for extensive data for model training.


This paper presents a DL-based framework tailored for LP-IoT channel estimation to tackle recent challenges in this field. The proposed framework emphasises simplicity and accuracy to facilitate easy deployment for channel estimation. It provides a new approach specifically designed to meet the requirements of resource-constraint LP-IoT systems. We developed two models, A and B, using real-time data from two scenarios with LP-IoT devices. The significant contributions of this research are outlined as follows:
\begin{itemize}
    \item Developed two advanced DL models for the estimation of wireless channels in low-power IoT networks, representing a substantial advancement in the realm of IoT communication.
    \item Designed the experimental testbed for collecting real-world data, utilising two LP-IoT devices, offering a practical and robust framework for empirical data collection and analysis.
    \item Presented a comparative analysis of our proposed technique against existing DL techniques and existing research, highlighting our model's strength and improvement. 
\end{itemize}

\section{System Model and Problem Formulation}\label{sec:SysModel}
This section describes the system model employed in this article for LP-IoT wireless channel estimation and articulates the research problem formulation in the subsequent subsections.

\subsection{System Model}
\label{subsec:SysModel}

\begin{figure*}[t]
    \centering
    \includegraphics[width=0.90\linewidth]{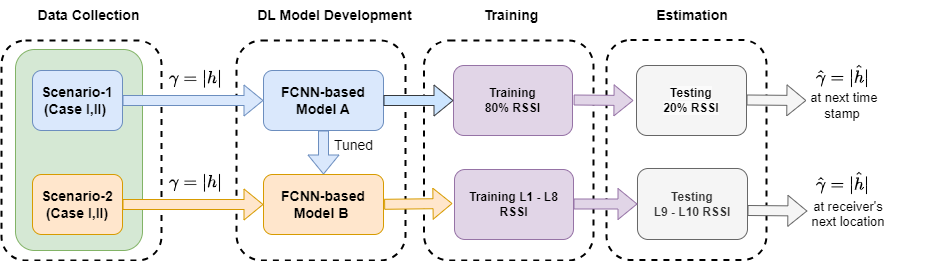}
    \caption{Visual representation of FCNN-based LP-IoT estimation model.}
    \label{fig:SysModel-AB}
\end{figure*}

The system model process commences with data collection across two distinct scenarios, each yielding RSSI data under varying conditions. Subsequently, two FCNN-based models, Model A and Model B, are constructed and tuned with the respective scenario datasets. Model A is trained using $80\%$ of the Scenario 1 RSSI data, while Model B utilises RSSI data from the first eight locations (L1-L8) in Scenario 2. The validation phase employs the remaining Scenario 1 data $(20\%)$ for Model A and the subsequent two locations (L9-L10) for Model B to test the model's performance in estimating the Lp-IoT wireless channel. This systematic approach exemplifies the application of DL in optimising signal strength prediction, which is crucial for enhancing LP-IoT wireless network performance. Figure \ref{fig:SysModel-AB} provides a high-level visual representation of our estimation model. Next, we discuss the core challenge of channel estimation in LP-IoT networks.

\subsection{Problem Formulation}
The standard wireless communication model in LP-IoT networks is given by $y=hx+n$, where $h$ denotes the wireless channel between LP-IoT transmitter and receiver, $x$ denotes the transmitted signal, and $n$ represents the noise in the system. Our objective is to estimate LP-IoT wireless channel gain in $dBm$, referred to as $RSSI$ and denoted as $\gamma=|h|$.
Building upon the fundamental understanding of the LP-IoT channel estimation challenge and the standard communication model, we now transition to a more focused exploration of our specifically designed solution in the subsequent section.

\begin{figure*}[t]
    \centering
    \includegraphics[width=0.98\linewidth]{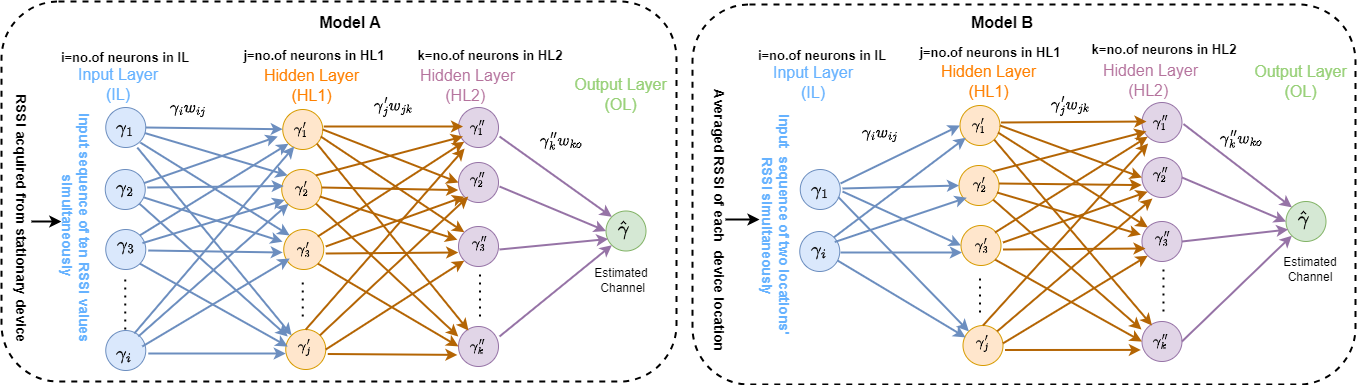}
    \caption{Visual representation of two FCNN-based estimation models using RSSI for LP-IoT networks.}
    \label{fig:Model-AB}
\end{figure*}

\section{DL Model Development}
\subsection{Model A} \label{Sec:ModelA}

For estimating wireless channels, specifically LP-IoT communication, we developed FCNN-based model A that is methodically segmented into four layers: an Input Layer (IL), two hidden layers (HL1 and HL2) and an Output Layer (OL). Each hidden layer contains $10$  neurons, employing the Leaky ReLU decision function in the connected layers of the neural network. Figure \ref{fig:Model-AB} illustrates the FCNN-based channel estimation model for LP-IoT devices, focusing on how it learns to map the input RSSI values to the corresponding output (estimated) RSSI values. A mathematical expression of FCNN-based channel estimation model A $(M_A)$ can be represented as follows:
\vspace{-1mm}
\begin{equation} \label{eq:Model-A}
    M_A \xrightarrow{\text{output}} \hat{\gamma}
\end{equation}
\begin{equation} \label{eq:Model-A0}
    \hat{\gamma} \longleftarrow  \delta_k(\gamma''_k) \longleftarrow \delta_j(\gamma'_j) \longleftarrow \gamma_i
\end{equation}


where $\gamma_{i}$, $ \gamma_{j}'$, $\gamma_{k}''$, $\hat{\gamma}$ denote the input from $IL$ to $HL1$, the output at $HL1$, output at $HL2$ and the final estimated output (single neuron) at $OL$ respectively. Furthermore, the variables $i$, $j$, and $k$ represent the number of neurons at $IL$, $HL1$, and $HL2$, respectively. The $\delta_j$ and $\delta_k$ represent the Leaky ReLU decision function, used in the connected layers of neural networks to introduce non-linearity. The Leaky ReLU function is a modification of the ReLU function that addresses the issue of dying ReLU. It includes a small constant value of $0.01$ in the ReLU function with the $\gamma$ and can be expressed as $\delta(\gamma) = max(0.01\gamma, \gamma)$. The expanded form of the parameters, used in equation \ref{eq:Model-A} are defined in the subsequent equations as follows:
\vspace{-3mm}
\begin{equation}
    \hspace{7mm}\gamma_i=\gamma_i w_{ij},\hspace{5mm} \text{where} \hspace{5mm}i=(1,\hdots10)\label{eq:IL}
\end{equation}
\vspace{-4mm}
\begin{equation} \label{eq:HL1}
    \gamma_{j}'=\delta_{j}(\gamma'_j)=\delta_{j} \left(\sum_{j=1}^{10} (\gamma'_j w_{ij}) \right), 
\end{equation}
\vspace{-2mm}
\begin{equation} \label{eq:HL2}
    \gamma_{k}''=\delta_{k}(\gamma''_k)=\delta_k \left(\sum_{k=1}^{10} (\gamma''_k w_{jk})\right), 
\end{equation}
\vspace{-2mm}
\begin{equation} \label{eq:OL}
    \hat{\gamma}= \sum_{o=1}^{1} (\gamma_k'' w_{ko}), 
\end{equation}
where $w_{ij}$, $w_{jk}$, $w_{ko}$ denote the synaptic weights from $IL$ to $HL1$, $HL1$ to $HL2$, and $HL2$ to $OL$ respectively. In our model's architecture, the value of variables $i$, $j$, and $k$ are set to 10 as we have $10$ neurons at each layer except the $OL$, and can be represented as $i=j=k=(1,2,3,\hdots, 10)$. We configure the $OL$ to generate a single estimated output for each given instance; therefore, we set the variable $o$ as $1$ to signify the singular neuron in the $OL$.

\subsubsection{Model Training}
The above-explained FCNN-based model structure has been designed for forward propagation, which does not provide the ideal estimation and provides higher estimation errors. To minimise these errors, we backpropagate by updating the weights in the designed FCNN-based estimation model. We employed the NAdam optimiser to update the weights and minimise the estimation error. Nadam is an enhanced version of the Adam algorithm that integrates Nesterov Momentum, resulting in improved performance compared to the ADAM algorithm \cite{Zaheer2019}. The NAdam optimiser can be expressed mathematically as:
\vspace{-4mm}
\begin{equation}
w_{t+1} = w_t - \frac{\eta}{\sqrt{\hat{v}_t} + \epsilon} \left(\beta_1 \hat{m}_t + \frac{(1 - \beta_1) g_t}{1 - \beta_1^t}\right),
\end{equation}
where $w_t$ represents the parameter at time step $t$, $\eta$ denotes the learning rate, $\hat{m}_t$ and $\hat{v}_t$ are bias-corrected (first and second-moment) estimates of the gradients, $g_t$ is the gradient, $\beta_1$ is the momentum decay rate, and $\epsilon$ represents a small constant used to prevent division by zero.
We trained the model until it reached the minimum estimation error. For this, we set the number of training epochs to $200$. The detailed outcome of this model has been discussed in Section \ref{sec:Results}. The collected RSSI data is treated as a time series, and the FCNN model has been trained by dividing the data into tiny sequences of data points. Each sequence is considered as a mini-list of data. The FCNN model A selects the first $10$ data points from the dataset and arranges them in a sequence of $10$. Additionally, it retrieves the next measured channel value and stores it as the target. The model takes a sequence, predicts the next data point, compares it with the target, and computes the error.
Likewise, the sequence window moves to the next value, takes $10$ inputs as a sequence and compares it with the next value (target). This process persists until all the data has been traversed. The MSE and Root Mean Squared Error (RMSE) evaluated the model's performance. The selected number of sequences can be altered. Nevertheless, the model provided the most optimal outcome by setting the data sequence as 10.
\begin{figure*}[t]
\centering
    \begin{subfigure}[t]{0.48\textwidth}
    \centering
    \includegraphics[width=\textwidth]{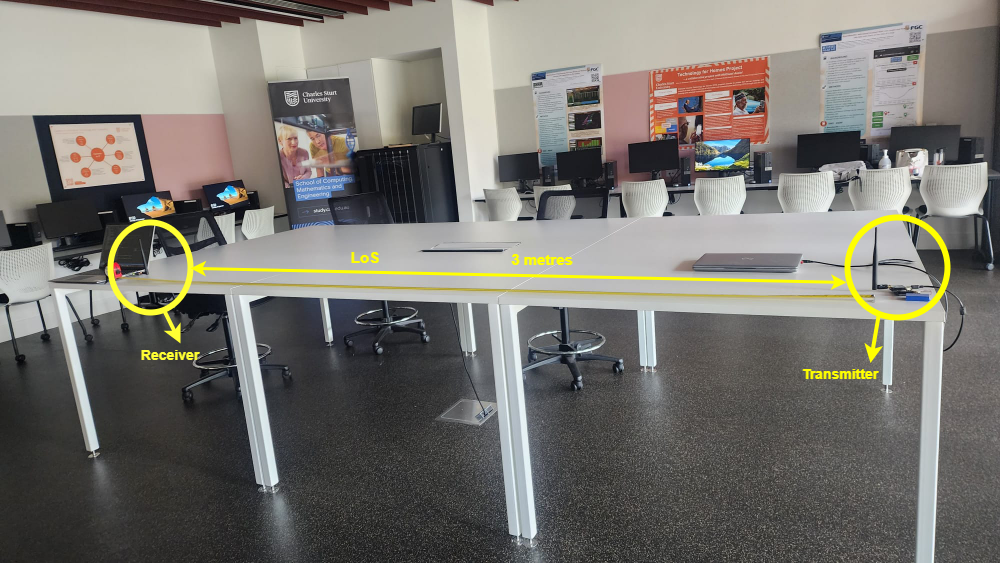}
    \caption{Case I - Line of Sight (LoS) environmental condition.} \label{Fig:S1C1}
\end{subfigure}\hfill
\begin{subfigure}[t]{0.48\textwidth}
    \centering
    \includegraphics[width=\textwidth]{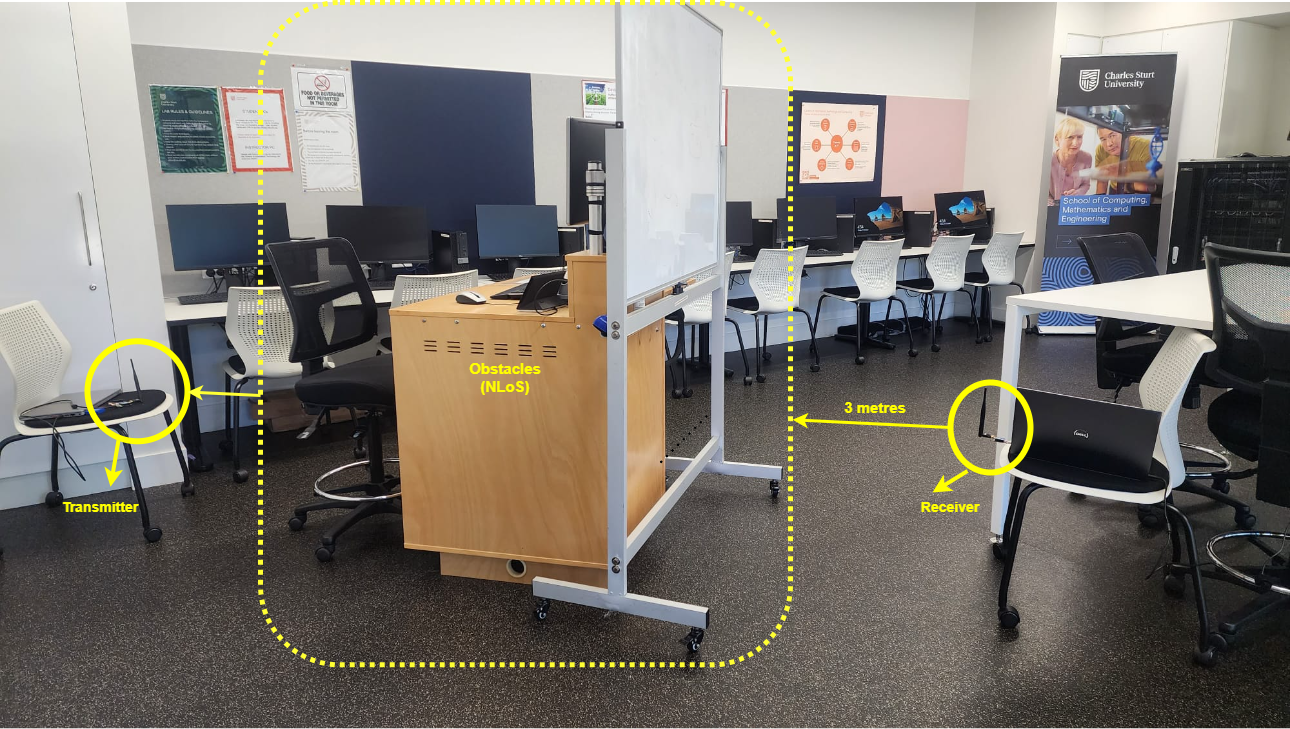}
    \caption{Case II - Non Line of Sight (NLoS) environmental condition.}   \label{Fig:S1C2}  
\end{subfigure}\hfill
\caption{Experimental setup for scenario 1-Fixed distance between transmitter and receiver.}\label{Fig:S1}
\end{figure*}

\begin{figure*}[!h]
\centering
    \begin{subfigure}[t]{0.48\textwidth}
    \centering
    \includegraphics[width=\textwidth]{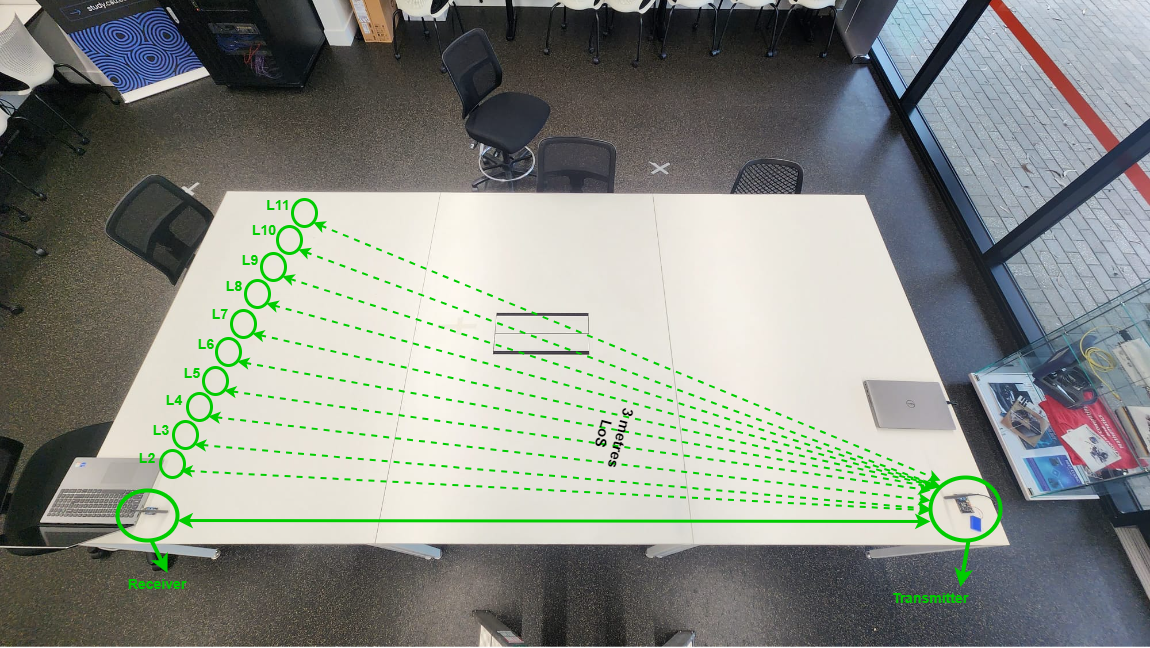}
    \caption{Case I - Line of Sight (LoS) environmental condition.} \label{Fig:S2C1}
\end{subfigure}\hfill
\begin{subfigure}[t]{0.48\textwidth}
    \centering
    \includegraphics[width=\textwidth]{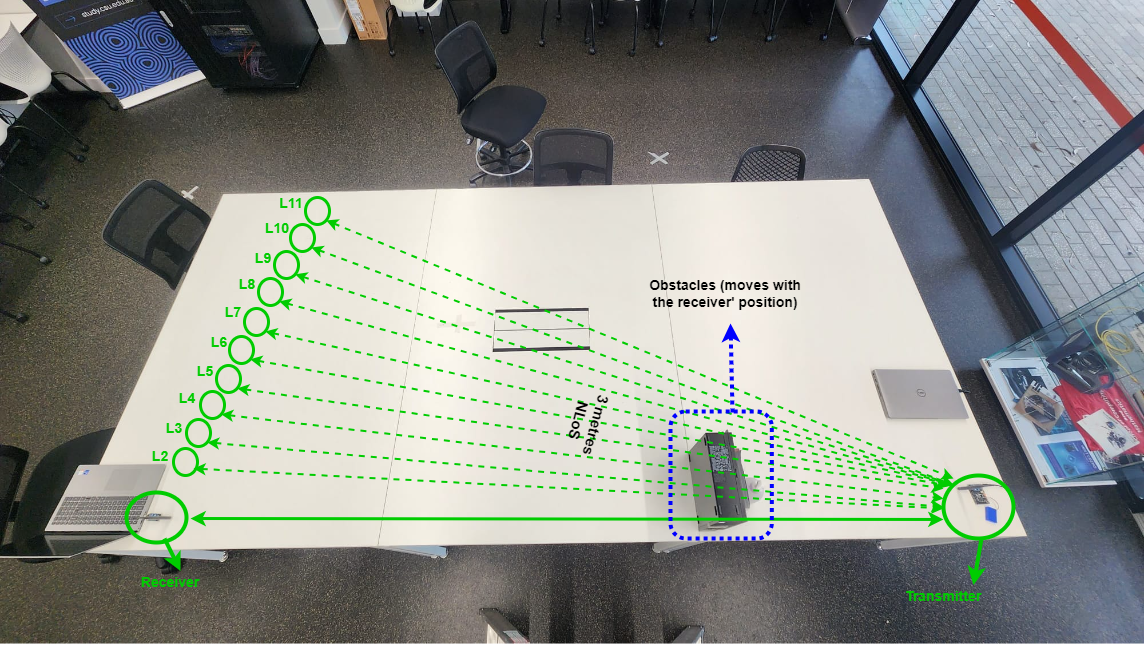}
    \caption{Case II - Non Line of Sight (NLoS) environmental condition.}   \label{fig:S2C2}  
\end{subfigure}\hfill
\caption{Experimental setup for scenario 2-Varying locations of the receiver at a fixed distance from the transmitter.}\label{Fig:S2}
\end{figure*}

\subsection{FCNN-based Model B} \label{Sec:ModelB}
For model B, we tuned model A for experiment scenario 2, in which we changed the receiver's location. We collected $250$ samples at each location and took the average of RSSI for each location. So, we have a single averaged RSSI for each location. We utilised the single RSSI for each location and predicted the RSSI for the next location at the next time stamp. For this, the sequence of the input neurons is now reduced from ten input neurons to two. We have utilised the two hidden layers and single output neurons to predict RSSI at the next location. The decision function and parameters optimiser are the same as in model A, Leaky ReLU and NAdam. The number of training epochs run to train the model was $300$ to reach the minimum MSE. The mathematical expression for model B can be defined as the same as in equations \ref{eq:Model-A} and \ref{eq:Model-A0}. For model B, the altered parameters in equation \ref{eq:Model-A0} are $\gamma_i$ at $IL$, and $\gamma'_j$ at $HL1$, and can be mathematically expressed as $\gamma_i=\gamma_i w_{ij}$, and $
    \gamma_{j}'=\delta_{j}(\gamma'_j)=\delta_{j} \left(\sum_{j=1}^{10} (\gamma'_j w_{ij}) \right)$, 
where $i=(1,2)$, and $j=(1,2,3,\hdots, 10)$.

\begin{figure*}[t]
\centering
    \begin{subfigure}[t]{0.5\textwidth}
    \centering
    \includegraphics[width=\textwidth]{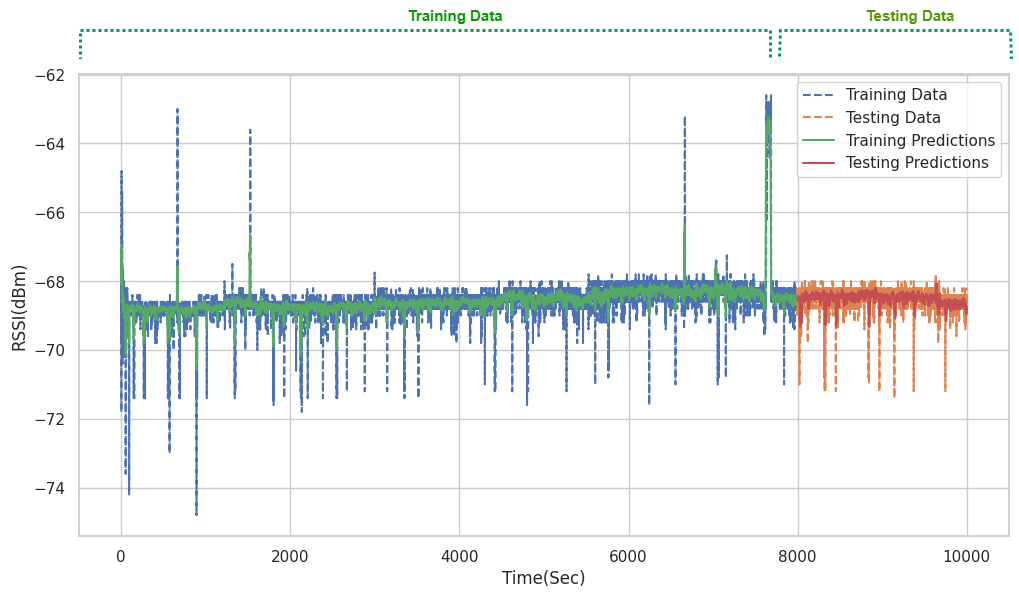}
    \caption{Scenario 1 - Case I: Training and testing phase performance.} \label{Fig:ModelA-Resuts-LoS}
\end{subfigure}\hfill
\begin{subfigure}[t]{0.5\textwidth}
    \centering
    \includegraphics[width=\textwidth]{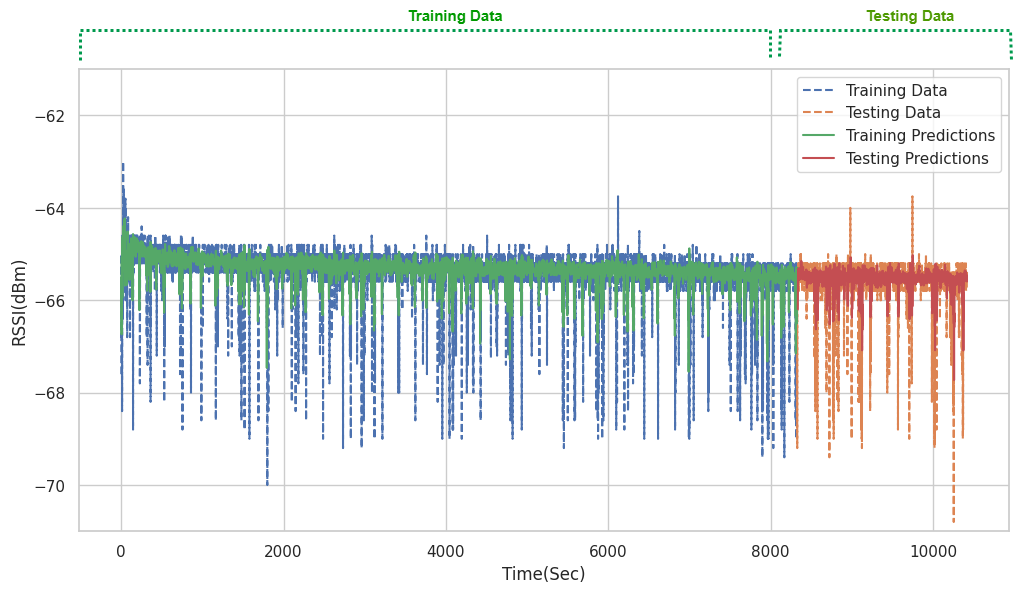}
    \caption{Scenario 1 - Case II: Training and testing phase performance.}   \label{Fig:ModelA-Results-NLoS}  
\end{subfigure}\hfill
\caption{FCNN-based Model A estimating RSSI at the next time stamp using historical RSSI from the LP-IoT stationary devices.}\label{Fig:ModelA-Results}
\end{figure*}

\begin{figure*}[t]
\centering
    \begin{subfigure}[t]{0.5\textwidth}
    \centering
    \includegraphics[width=\textwidth]{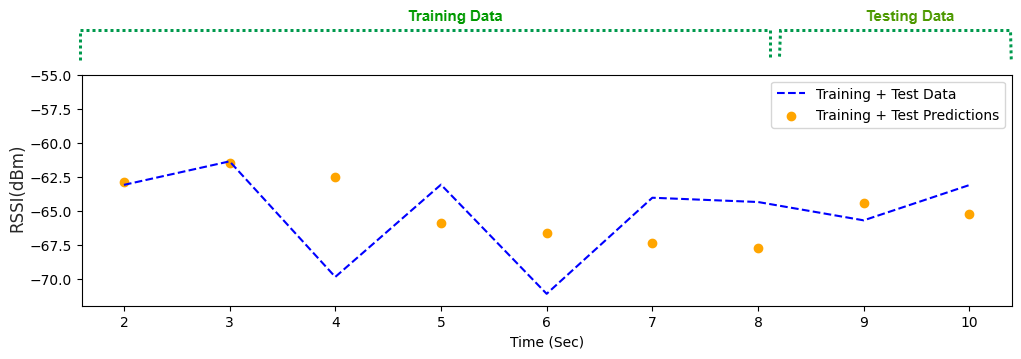}
    \caption{Scenario 2 - Case I: Training and testing phase performance.} \label{Fig:ModelB-Results-LoS}
\end{subfigure}\hfill
\begin{subfigure}[t]{0.5\textwidth}
    \centering
    \includegraphics[width=\textwidth]{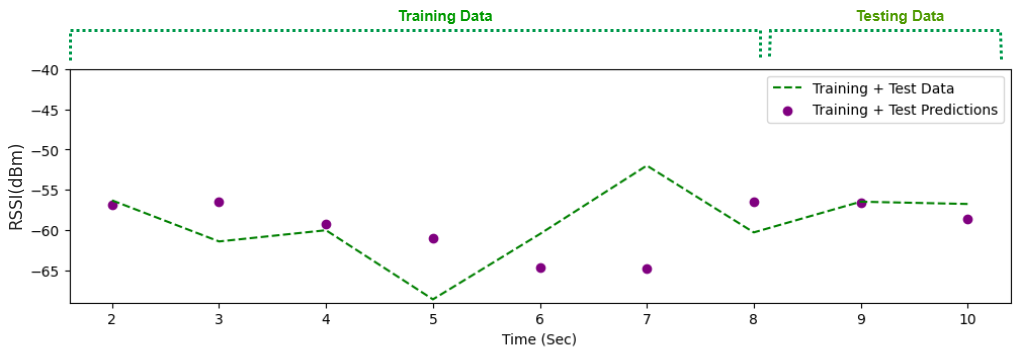}
    \caption{Scenario 2 - Case II: Training and testing phase performance.}   \label{Fig:ModelB-Results-NLoS}  
\end{subfigure}\hfill
\caption{FCNN-based Model B estimating RSSI at the next receiver locations of LP-IoT device utilising previous locations.}\label{Fig:ModelB-Results}
\end{figure*}

\section{Experimental Data Collection}

To implement the developed LP-IoT wireless channel estimation model, we delve into the experimental framework established to assess the effectiveness of our suggested approaches in IoT communication. To accomplish this, we set up two LP-IoT devices, one serving as a transmitter and the other as a receiver within a controlled network laboratory setting. The setup employs two scenarios based on varying environmental conditions and the receiver's position. These scenarios are described in the subsequent subsections.


\subsection{Scenario 1: Stationary Devices with Fixed Distances}

Scenario 1 was designed to investigate the impact of wireless channels on stationary devices in an indoor environment. This exploration was fundamental in understanding the effect of indoor settings on wireless communication.

\subsubsection{Case I}

In case I of Scenario 1, the experimental setup was established on a centre table in the network laboratory. The transmitter and receiver were placed at a fixed distance of $3$ metres, ensuring a direct Line-of-Sight (LoS), as depicted in Figure \ref{Fig:S1C1}. The LoS setup was chosen to assess the baseline communication performance, providing an unobstructed transmission route in a controlled laboratory environment.

\subsubsection{Case II}
Case II of Scenario 1 was designed to simulate Non-Line-of-Sight (NLoS) conditions, introducing physical barriers to the transmission path. In this setup, the transmitter and receiver were strategically placed on opposite sides of a standard lecture table, as depicted in Figure \ref{Fig:S1C2}. The table was comprised of wood and was equipped with PowerPoint and a screen, which served as the obstructing medium. Also, common office objects such as a whiteboard and a chair were utilised as physical obstructions. These objects were strategically positioned to obstruct the direct transmission path between the transmitter and receiver, simulating a real-world NLoS environment. 

\titlespacing\subsection{0pt}{-2pt plus 8pt minus 2pt}{0pt plus 2pt minus 2pt}
\subsection{Scenario 2: Receiver's Location Variation Assessment}
The next scenario has been designed to explore the impact of varying the receiver's position within the laboratory environment, thereby assessing the influence of spatial dynamics on signal transmission. 

\subsubsection{Case I} 

In Case I of Scenario 2, the focus was on evaluating the LoS condition, emphasising the variability of receiver locations. The data was collected by placing the receiver at multiple points on a designated table within the laboratory, as shown in Figure \ref{Fig:S2C1}. These points were selected to represent a diverse range of spatial configurations, aiming to comprehensively assess how varying locations with fixed distances within a LoS condition affect signal quality and strength.

\titlespacing\subsubsection{0pt}{2pt plus -1pt minus 2pt}{0pt plus 2pt minus 2pt}
\subsubsection{Case II} 

In Case II of Scenario 2, we adjusted our experimental setup to introduce the NLoS condition, mirroring the approach used in Scenario 1. Some common laboratory items were placed as obstructions to create a realistic NLoS environment, as depicted in Figure \ref{fig:S2C2}. These included a small CPU, a book, and a hand sanitiser strategically placed within the testing area. The data was then collected from this zone by slightly varying the receiver's position from its previous position, allowing for the analysis across $11$ different locations. This setup provided valuable insights into the challenges and behaviour of signal transmission in NLoS conditions, which are commonly encountered in practical indoor communication scenarios.

\subsection{Data Preprocessing}
For the development of Model A, the experiment was structured to include five separate readings, 
each consisting of 10,000 samples. The five readings were averaged for each time step, yielding a consolidated dataset. This method of averaging served to smooth out any anomalies or outliers that might have occurred in individual readings, thereby ensuring a higher degree of accuracy and consistency in the results. Crucially, this process was carried out for LoS and NLoS conditions, resulting in two sets of 10,000 averaged samples each.

For the establishment of Model B, the experiment was conducted across 11 distinct locations, each chosen to represent a diverse range of environmental conditions typical in IoT network deployments. At each of these locations, 250 individual RSSI samples were meticulously collected. By averaging the 250 samples at each location, we distilled the data into a singular, representative RSSI value for that specific location. Each of these readings served as the main RSSI value for the corresponding location within our model.

\input{Tables/Comparison-Traditional-Final}
\input{Tables/Comparison-DL-Final}

\section{Results and Discussion} \label{sec:Results}

\subsection{Estimation}
This section explains the outcome acquired from the developed FCNN-based models. The model performance was evaluated using the LoS and NLoS conditions. For both conditions, we trained the 80\% RSSI data points, which is trained on MSE $0.17\hspace{1mm}dBm$ for LoS and $0.24\hspace{1mm}dBm$ for NLoS, whereas the testing results provide the MSE of $0.11\hspace{1mm}dBm$ for LoS and $0.23\hspace{1mm}dBm$ for NLoS as seen in Figures \ref{Fig:ModelA-Resuts-LoS} and \ref{Fig:ModelA-Results-NLoS}. From the results, we observe that our model A behaves well in both conditions, providing robustness to our model. Next, we tuned model A to develop model B by reducing the number of sequences as explained in Section \ref{Sec:ModelB}. We evaluated model B's training and testing results using LoS and NLoS conditions. The training MSE of LoS and NLoS conditions is $15.00\hspace{1mm}dBm$ and $39.35\hspace{1mm}dBm$, whereas the testing MSE of $2.63\hspace{1mm}dBm$ and $4.82\hspace{1mm}dBm$ for LoS and NLoS conditions, respectively. The graphical results of model B in NLoS and NLoS condition can be seen in Figures \ref{Fig:ModelB-Results-LoS} and \ref{Fig:ModelB-Results-NLoS}. As model B is the tuned version of model A, we analyse model A for deeper analysis by comparing its results with existing literature and DL-based techniques in the subsequent subsection.

\subsection{Analysis of Results}
To compare our model with other DL-based techniques, such as Recurrent Neural Network (RNN) and Long Short-Term Memory (LSTM), we employed the identical dataset utilised in our proposed approach. These techniques require tuning a few parameters, such as the number of neurons and training epochs, to reach the minimum error. The numeric results of these techniques applied to our data can be found in Table \ref{tab:Comparison}. Upon in-depth consideration, we can see that the FCNN-based proposed model A provides slightly lower training MSE and RMSE than RNN and LSTM. This model's testing MSE and RMSE is approximately the same as RNN and LSTM for Case I. However, it outperforms these methods in Case II, which shows that our model performs well in a dynamic environment. Also, the major factors to observe are that the training time, memory usage and number of trainable parameters of RNN and LSTM are quite higher than the proposed model, as mentioned in Table \ref{tab:Comparison}, which shows the simplicity of our model.

Finally, we compare our proposed models with the existing research, exhibiting notable improvements in performance metrics over existing models in the literature. In \cite{Raj2021}, the author proposed a model that reported an average testing MSE across all datasets of $41.92\hspace{1mm}dBm$ in indoor RSSI prediction. In comparison, our models A and B achieved a testing MSE of $0.41\hspace{1mm}dBm$ and $3.72\hspace{1mm}dBm$ (averaged across two cases of scenario $1$ and $2$), which is significantly lower. These results demonstrate a substantial improvement in the FCNN-based channel estimation model. Although our developed models perform well for the given scenarios, they can be expanded to encompass more complex scenarios, such as varying distances between the LP-IoT devices. This perspective offers an exciting avenue for further development, enhancing the model's applicability in diverse IoT environments.
\section{Conclusion and Future Work}
In conclusion, this paper represents a significant advancement in the realm of LP-IoT wireless channel estimation. By analysing the existing challenges in channel estimation for LP-IoT networks, we have developed two wireless channel estimation models that stand out in accuracy and efficiency. A key highlight of our research is the significant reduction in estimation error. The average estimation error in the existing literature is $41.92 \hspace{1mm}dBm$. However, our models A and B perform significantly better with average MSE of $0.41\hspace{1mm}dBm$ and $3.72\hspace{1mm}dBm$ in Scenario $1$ and $2$, respectively.
Furthermore, these models demonstrate a marked advantage in computational efficiency. They require fewer training parameters and utilise less memory than other DL-based techniques, making them ideal for LP-IoT applications. As a part of our future work, we plan to perform a detailed complexity analysis of our model by practically deploying these models on LP-IoT devices. We also plan to enhance the model's adaptability to diverse environments, including multi-room and multi-floor settings, with multi-antenna devices.

 \nocite{*} 
\bibliographystyle{unsrt}
\bibliography{RefList.bib}  

\end{document}

%% file: Tables/Comparison-Traditional-Final.tex
\renewcommand{\arraystretch}{1.3} 


%% file: Tables/Comparison-DL-Final.tex

\begin{table*}[t]
\centering
\tiny
\caption{The comparative analysis of FCNN-based estimation model A with other DL-based techniques.}
\label{tab:Comparison}

\begin{tabular}{|lllllllllc|}
\hline
\rowcolor[HTML]{ECF4FF} 
\multicolumn{1}{|c|}{\cellcolor[HTML]{ECF4FF}} &
  \multicolumn{2}{l|}{\cellcolor[HTML]{ECF4FF}\textbf{Training}} &
  \multicolumn{2}{l|}{\cellcolor[HTML]{ECF4FF}\textbf{Testing}} &
  \multicolumn{1}{c|}{\cellcolor[HTML]{EFEFEF}} &
  \multicolumn{1}{c|}{\cellcolor[HTML]{EFEFEF}} &
  \multicolumn{1}{c|}{\cellcolor[HTML]{EFEFEF}} &
  \multicolumn{1}{c|}{\cellcolor[HTML]{EFEFEF}} &
  \multicolumn{1}{l|}{\cellcolor[HTML]{EFEFEF}} \\ \cline{2-5}
\rowcolor[HTML]{ECF4FF} 
\multicolumn{1}{|c|}{\multirow{-2}{*}{\cellcolor[HTML]{ECF4FF}\textbf{\begin{tabular}[c]{@{}c@{}}Estimation\\ Errors\end{tabular}}}} &
  \multicolumn{1}{l|}{\cellcolor[HTML]{ECF4FF}\textbf{MSE}} &
  \multicolumn{1}{l|}{\cellcolor[HTML]{ECF4FF}\textbf{RMSE}} &
  \multicolumn{1}{l|}{\cellcolor[HTML]{ECF4FF}\textbf{MSE}} &
  \multicolumn{1}{l|}{\cellcolor[HTML]{ECF4FF}\textbf{RMSE}} &
  \multicolumn{1}{c|}{\multirow{-2}{*}{\cellcolor[HTML]{EFEFEF}\textbf{\begin{tabular}[c]{@{}c@{}}Training \\ Time (Secs)\end{tabular}}}} &
  \multicolumn{1}{c|}{\multirow{-2}{*}{\cellcolor[HTML]{EFEFEF}\textbf{\begin{tabular}[c]{@{}c@{}}Testing\\  Time (Secs)\end{tabular}}}} &
  \multicolumn{1}{c|}{\multirow{-2}{*}{\cellcolor[HTML]{EFEFEF}\textbf{\begin{tabular}[c]{@{}c@{}}Training Memory\\  Usage (KBs)\end{tabular}}}} &
  \multicolumn{1}{c|}{\multirow{-2}{*}{\cellcolor[HTML]{EFEFEF}\textbf{\begin{tabular}[c]{@{}c@{}}Testing Memory\\  Usage (KBs)\end{tabular}}}} &
  \multicolumn{1}{l|}{\multirow{-2}{*}{\cellcolor[HTML]{EFEFEF}\textbf{\begin{tabular}[c]{@{}l@{}}No. of Trainable\\  Parameaers\end{tabular}}}} \\ \hline
\rowcolor[HTML]{CBCEFB} 
\multicolumn{10}{|c|}{\cellcolor[HTML]{CBCEFB}\textbf{Proposed Fully Connected Neural Network (FCNN)-based Model A}} \\ \hline
\multicolumn{1}{|l|}{\textbf{\begin{tabular}[c]{@{}l@{}}Scenario 1\\ Case I\end{tabular}}} &
  \multicolumn{1}{l|}{0.17} &
  \multicolumn{1}{l|}{0.41} &
  \multicolumn{1}{l|}{0.11} &
  \multicolumn{1}{l|}{0.34} &
  \multicolumn{1}{l|}{0.954} &
  \multicolumn{1}{l|}{0.006} &
  \multicolumn{1}{l|}{675668} &
  \multicolumn{1}{l|}{676296} &
   \\ \cline{1-9}
\multicolumn{1}{|l|}{\textbf{\begin{tabular}[c]{@{}l@{}}Scenario 1\\ Case II\end{tabular}}} &
  \multicolumn{1}{l|}{0.24} &
  \multicolumn{1}{l|}{0.48} &
  \multicolumn{1}{l|}{0.23} &
  \multicolumn{1}{l|}{0.48} &
  \multicolumn{1}{l|}{1.792} &
  \multicolumn{1}{l|}{0.002} &
  \multicolumn{1}{l|}{656028} &
  \multicolumn{1}{l|}{656272} &
  \multirow{-2}{*}{231} \\ \hline
\rowcolor[HTML]{DAE8FC} 
\multicolumn{10}{|c|}{\cellcolor[HTML]{DAE8FC}\textbf{Recurrent Neural Network (RNN)-based Model}} \\ \hline
\multicolumn{1}{|l|}{\textbf{\begin{tabular}[c]{@{}l@{}}Scenario 1\\ Case I\end{tabular}}} &
  \multicolumn{1}{l|}{0.37} &
  \multicolumn{1}{l|}{0.61} &
  \multicolumn{1}{l|}{0.10} &
  \multicolumn{1}{l|}{0.32} &
  \multicolumn{1}{l|}{300.687} &
  \multicolumn{1}{l|}{0.008} &
  \multicolumn{1}{l|}{940784} &
  \multicolumn{1}{l|}{706372} &
   \\ \cline{1-9}
\multicolumn{1}{|l|}{\textbf{\begin{tabular}[c]{@{}l@{}}Scenario 1\\ Case II\end{tabular}}} &
  \multicolumn{1}{l|}{0.22} &
  \multicolumn{1}{l|}{0.47} &
  \multicolumn{1}{l|}{0.24} &
  \multicolumn{1}{l|}{0.49} &
  \multicolumn{1}{l|}{314.708} &
  \multicolumn{1}{l|}{0.006} &
  \multicolumn{1}{l|}{975988} &
  \multicolumn{1}{l|}{721780} &
  \multirow{-2}{*}{4353} \\ \hline
\rowcolor[HTML]{D8F4D7} 
\multicolumn{10}{|c|}{\cellcolor[HTML]{D8F4D7}\textbf{Long Short Term Memory (LSTM)-based Model}} \\ \hline
\multicolumn{1}{|l|}{\textbf{\begin{tabular}[c]{@{}l@{}}Scenario 1\\ Case I\end{tabular}}} &
  \multicolumn{1}{l|}{0.37} &
  \multicolumn{1}{l|}{0.61} &
  \multicolumn{1}{l|}{0.11} &
  \multicolumn{1}{l|}{0.32} &
  \multicolumn{1}{l|}{292.700} &
  \multicolumn{1}{l|}{0.230} &
  \multicolumn{1}{l|}{975988} &
  \multicolumn{1}{l|}{727884} &
   \\ \cline{1-9}
\multicolumn{1}{|l|}{\textbf{\begin{tabular}[c]{@{}l@{}}Scenario 1\\ Case II\end{tabular}}} &
  \multicolumn{1}{l|}{0.22} &
  \multicolumn{1}{l|}{0.47} &
  \multicolumn{1}{l|}{0.24} &
  \multicolumn{1}{l|}{0.49} &
  \multicolumn{1}{l|}{312.110} &
  \multicolumn{1}{l|}{0.160} &
  \multicolumn{1}{l|}{1021652} &
  \multicolumn{1}{l|}{742744} &
  \multirow{-2}{*}{26641} \\ \hline
\end{tabular}
\end{table*}